\documentclass{article}
\usepackage{graphicx} %

\usepackage{cite}
\usepackage{amsmath,amssymb,amsfonts}
\usepackage{algorithmic}
\usepackage{algorithm}
\usepackage{graphicx}
\usepackage{textcomp}
\usepackage{xcolor}

\usepackage[left=2cm,right=2cm,top=2.5cm,bottom=2.5cm]{geometry} %

\usepackage{tabulary}
\usepackage{multirow}

\usepackage{soul}

\usepackage{adjustbox}

\usepackage{bbm}
\usepackage{amsthm}

\usepackage{xspace}

\newcommand{\textmacro}[2]{\newcommand{#1}{#2\xspace}}

\newcommand{\eg}{{\it e.g.,~}}
\newcommand{\ie}{{\it i.e.,~}}

\newcommand{\etal}{{\em ~et al.}}
\textmacro{\rn}{ResNet152}
\textmacro{\rnbase}{ResNet\_1\_1\_1\_1}
\textmacro{\rnfull}{ResNet\_3\_8\_36\_3}

\usepackage{fancyhdr}
\title{RRR-Net: Reusing, Reducing, and Recycling a Deep Backbone Network}
\author{Haozhe Sun $^{1}$ \and Isabelle Guyon $^{1,2}$ \and Felix Mohr $^{3}$ \and Hedi Tabia $^{4}$}
\date{$^{1}$ \quad LISN/CNRS/INRIA, Université Paris-Saclay, France\\
$^{2}$ \quad ChaLearn, USA \\
$^{3}$ \quad Universidad de La Sabana, Colombia\\
$^{4}$ \quad IBISC, Univ. Evry, Université Paris-Saclay, France \\ 
Email: haozhe.sun@universite-paris-saclay.fr}

\begin{document}

\maketitle

\begin{abstract}

It has become mainstream in computer vision and other machine learning domains to reuse backbone networks pre-trained on large datasets as preprocessors. Typically, the last layer is replaced by a shallow learning machine of sorts; the newly-added classification head and (optionally) deeper layers are fine-tuned on a new task. Due to its strong performance and simplicity, a common pre-trained backbone network is \rn.
However, \rn is relatively large and induces inference latency. In many cases, a compact and efficient backbone with similar performance would be preferable over a larger, slower one. This paper investigates techniques to reuse a pre-trained backbone with the objective of creating a smaller and faster model. Starting from a large \rn backbone pre-trained on ImageNet, we first reduce it from 51 blocks to 5 blocks, reducing its number of parameters and FLOPs by more than 6 times, without significant performance degradation. Then, we split the model after 3 blocks into several branches, while preserving the same number of parameters and FLOPs, to create an ensemble of sub-networks to improve performance. Our experiments on a large benchmark of $40$ image classification datasets from various domains suggest that our techniques match the performance (if not better) of ``classical backbone fine-tuning'' while achieving a smaller model size and faster inference speed.  

\end{abstract}

\section{Background and motivations}

Over the last decade, Deep Learning has set new standards in computer vision. Tasks in this area include the recognition of street signs, placards, and living beings. While it has achieved state-of-the-art in various academic and industrial fields, training deep networks from scratch requires massive amounts of data and hours of GPU training, which prevents it from being deployed in data-scarce and resource-scarce scenarios.

This limitation has been mainly addressed through the notion of {\em Transfer learning} \cite{panSurveyTransferLearning2010}.
Here, knowledge is transferred from a {\em source domain} (typically learned from a large dataset) to one or several {\em target domains} (typically with less available data).
A common transfer learning approach is last-layer fine-tuning~\cite{hintonReducingDimensionalityData2006}, in which a considerable part (the \emph{backbone}) of a pre-trained deep network is reused; only the last layer is replaced with a new classifier and trained to the new task at hand. Depending on the distribution shift between the source domain and target domains, more layers may be fine-tuned. Pre-trained networks that have been used for fine-tuning range from the historical AlexNet~\cite{krizhevskyImageNetClassificationDeep2012} to various ResNets~\cite{heDeepResidualLearning2016, zagoruykoWideResidualNetworks2016}.

Modern neural networks are thought of as being ``the bigger, the better'' as big networks keep beating large benchmarks (such as ImageNet~\cite{russakovskyImageNetLargeScale2015}). However, they are considerably over-parameterized when applied to smaller tasks. There is evidence that low-complexity models can, in some conditions, lead to comparably good or better performance~\cite{brigatoCloseLookDeep2021}. Our charter is to elaborate on the basic ``Reuse'' methodology described above (replacing the last layer with a new classifier) by applying the three classical resource-saving ``precepts'' to the greatest possible extent: ``\textbf{R}euse, \textbf{R}educe, and \textbf{R}ecycle''~\cite{kochReducedReusedRecycled2021}. For simplicity of demonstration, we limit ourselves to the popular \rn model~\cite{heDeepResidualLearning2016}, which provides us with enough flexibility to carry out our analyses. In this architecture, the ``Reduce'' step means to reduce the number of residual blocks in the network, and ``Recycle'' is done by creating a voting ensemble from existing layers. We leverage the reusability of a pre-trained backbone in part.

Our experimental results show that, for many tasks, a pre-trained \rn can be significantly reduced in order to benefit from faster inference speed while achieving similar (if not better) performance. Our evaluation is quite extensive, being based on $40$ datasets from various domains~\cite{ullahMetaalbumMultidomainMetadataset2022}. The positive results obtained with the \rn model suggest generalizing our techniques to other architectures.

\section{Related work}

This paper combines several ideas: {\em reusing} a pre-trained backbone network, {\em reducing} it, and splitting the model into several branches to use ensemble techniques ({\em recycling}). We briefly review related work in these three domains.

Given the availability of neural networks trained on large datasets, the vanilla fine-tuning approach~\cite{hintonReducingDimensionalityData2006} remains the method of choice for transfer learning with neural networks. Guo\etal~\cite{guoSpottuneTransferLearning2019} propose to adaptively fine-tune pre-trained models on a per-instance basis. Wortsman\etal~\cite{wortsmanModelSoupsAveraging2022} and Liu\etal~\cite{liuKnowledgeFlowImprove2019} propose methods to merge knowledge from multiple pre-trained models. Our method can be seen as performing neural architecture searches in the search space defined by ResNet152~\cite{heDeepResidualLearning2016}, which allows {\it reusing} its pre-trained parameters.

\noindent\textbf{Model pruning} is one way to reduce the storage/computing resource requirements of neural networks. It removes redundant parts (parameters, channels, etc.) that do not significantly contribute to the performance. Model pruning can happen at different granularities, depending on the topology constraints of the removed parts~\cite{lecunOptimalBrainDamage1989, hanLearningBothWeights2015, wenLearningStructuredSparsity2016, liuLearningEfficientConvolutional2017, liPruningFiltersEfficient2017}. The {\em reduction} step of our methodology (Section~\ref{sec:reduction_step}) can be categorized as a coarse-grained (block-wise) pruning approach, as opposed to finer-grained (element/channel-wise) pruning approaches. We compare our {\em reduction} step and finer-grained pruning approaches in Section~\ref{sec:pruning_granularity_study}.

\noindent\textbf{Ensemble methods}~\cite{juRelativePerformanceEnsemble2018} combine multiple models to improve generalization and robustness. Neural network ensemble methods include bagging~\cite{breimanBaggingPredictors1996}, Snapshot Ensemble~\cite{huangSnapshotEnsemblesTrain2017}, and Fast Geometric Ensemble~\cite{garipovLossSurfacesMode2018}. Several works~\cite{dingTrunkbranchEnsembleConvolutional2017, kimSplitNetLearningSemantically2017, opitzEfficientModelAveraging2016, malekhosseiniSplittingConvolutionalNeural2020} investigate tree-structured neural networks. The regularization technique dropout~\cite{tompsonEfficientObjectLocalization2015} can also be interpreted as an implicit ensemble technique. We use ensembling to {\it recycle} parts of the pre-trained network in Section~\ref{sec:recycling}.

\newcommand{\blocka}[2]{\multirow{3}{*}{\(\left[\begin{array}{c}\texttt{3$\times$3, #1}\\[-.1em] \texttt{3$\times$3, #1} \end{array}\right]\)$\times$#2}
}

\newcommand{\blockb}[3]{\multirow{3}{*}{\(\left[\begin{array}{c}\texttt{$1 \times 1$, #2}\\[-.1em] \texttt{$3 \times 3$, #2}\\[-.1em] \texttt{$1 \times 1$, #1}\end{array}\right]\)$\times$#3}
}

\renewcommand\arraystretch{1.1}
\setlength{\tabcolsep}{3pt}

\begin{figure*}[t]
\begin{center}
\resizebox{\linewidth}{!}{
\begin{tabular}{c|c|c|c|c|c|c|c}
\hline

Phases & output size & \multicolumn{2}{c}{Blocks} \vline &  \multicolumn{2}{c}{Num. parameters} \vline & \multicolumn{2}{c}{FLOPs ($10^6$)}  \\
\hline

 &  & Baseline & Ours (conv4\_x, conv5\_x: each branch) & Baseline & Ours & Baseline & Ours   \\
\hline

\multirow{2}{*}{conv1} & \multirow{2}{*}{$32\times 32$} & 7$\times $7, 64, conv (stride 2) & 7$\times $7, 64, conv (stride 2) & \multirow{2}{*}{9.5k} & \multirow{2}{*}{9.5k} & \multirow{2}{*}{40} & \multirow{2}{*}{40}   \\
  &  & 3$\times $3, max-pool (stride 2)  & 3$\times $3, max-pool (stride 2)  &  &  & \\
\hline

\multirow{3}{*}{conv2\_x} & \multirow{3}{*}{$32\times 32$} & \blockb{$256$}{$64$}{3} & \blockb{$256$}{$64$}{1} & \multirow{3}{*}{70k (75k)} & \multirow{3}{*}{70k (75k)} & \multirow{3}{*}{73 (78)} & \multirow{3}{*}{73 (78)}  \\
  &  &  &  &  &  &  \\
  &  &  &  &  &  &  \\
\hline

\multirow{3}{*}{conv3\_x} & \multirow{3}{*}{$16\times 16$} & \blockb{$512$}{$128$}{8} & \blockb{$512$}{$128$}{1} & \multirow{3}{*}{280k (379k)} & \multirow{3}{*}{280k (379k)} & \multirow{3}{*}{72 (123)} & \multirow{3}{*}{72 (123)}  \\
  &  &  &  &  &  &  \\
  &  &  &  &  &  &  \\
\hline

\multirow{3}{*}{conv4\_x} & \multirow{3}{*}{$8\times 8$} & \blockb{$1024$}{$256$}{36} & \blockb{$(256 / \sqrt{b} - a) \times 4$}{$256 / \sqrt{b} - a$}{1} & \multirow{3}{*}{1.12M (1.51M)} & \multirow{3}{*}{136k (333k) $\times$ 8} & \multirow{3}{*}{72 (122)} & \multirow{3}{*}{9 (30) $\times$ 8}  \\
  &  &  &  &  &  &  \\
  &  &  &  &  &  &  \\
\hline

\multirow{3}{*}{conv5\_x} & \multirow{3}{*}{$4\times 4$} & \blockb{$2048$}{$512$}{3} & \blockb{$(512 / \sqrt{b} - a) \times 4$}{$512 / \sqrt{b} - a$}{1} & \multirow{3}{*}{4.46M (6.04M)} & \multirow{3}{*}{553k (745k) $\times$ 8} & \multirow{3}{*}{72 (122)} & \multirow{3}{*}{9 (15) $\times$ 8}   \\
  &  &  &  &  &  &  \\
  &  &  &  &  &  &  \\
\hline

Total &  &  &  & 58.23M & 9.09 M & 3799 & 601  \\
\hline

\hline
\end{tabular}
}
\end{center}

\caption{{\bf The structure and approximated statistics of models.} We call ``Baseline" the original ResNet152~\cite{heDeepResidualLearning2016}. We call ``Ours" the reduced and recycled architecture. Reduction means removing blocks from each phase (until one block in each phase). Recycling means splitting blocks in the conv4\_x and conv5\_x phases into multiple branches while preserving the total number of parameters and FLOPs. This table assumes the input size is 128$\times$ 128. Building blocks are shown in brackets, with the numbers of blocks stacked. For conv4\_x and conv5\_x in the ``Ours" column, blocks are split into branches, where $b$ denotes the number of branches, and $a$ is a scalar to adjust the model size. The ``Num. parameters" and ``FLOPs" columns assume $b=8$. The values in parentheses indicate the statistics of the first block of each phase, which is responsible for downsampling. k $=10^3$, M $=10^6$. FLOPs mean floating-point operations or multiply-adds; they are measured with respect to a single $128\times 128 \times 3$ image using the open source software Pytorch-OpCounter~\cite{Lyken17PytorchOpCounterCount}. } 
\label{fig:arch}

\end{figure*}

This paper introduces a method to enhance the compactness and efficiency of neural network architectures. It highlights the importance of considering the recycling of pre-trained models. The paper is not limited to merely proposing a model compression technique but also emphasizes how efficiently one could reuse existing knowledge in pre-trained models.

\section{The RRR Principle for Image Classification}
\label{sec:RRR_principle}

When creating an artifact based on the RRR principle~\cite{kochReducedReusedRecycled2021}, one selects an object of interest for \emph{reuse}, then \emph{reduces} its complexity, and finally \emph{recycles} parts of it for some innovative modification. Adopting the RRR principle, our suggestion to arrive at a high-quality neural network for image classification is as follows:

\begin{itemize}
    \item {\bf Reuse:} Due to its great performance and simplicity, we choose a \rn pre-trained on ImageNet~\cite{russakovskyImageNetLargeScale2015} as the basis (backbone).
    \item {\bf Reduction:} We reduce the backbone to its basic version by eliminating potentially dispensable blocks.
    \item {\bf Recycling:} We keep the first few blocks (stump) as the feature extractor and split the remaining blocks into multiple branches to create a voting ensemble.
\end{itemize}

We next describe each of these points in more detail.

\subsection{Reuse: Description of the pre-trained backbone}

\begin{figure}[t]
\begin{center}
\includegraphics[width=\linewidth]{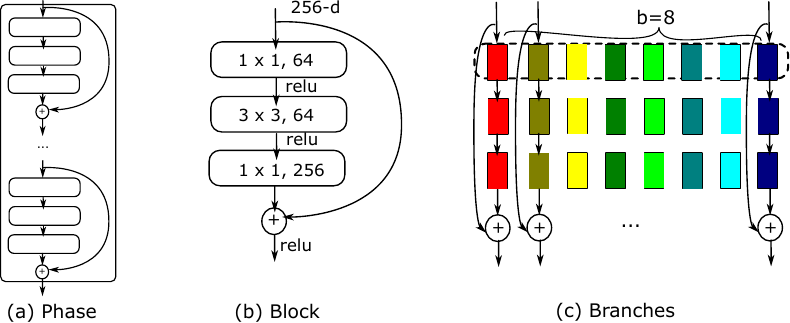}
\end{center}
\caption{Illustration of phase, block, and branches. (a) A phase represents a group of blocks. (b) A block, also called bottleneck block, stacks 3 convolutional layers with a skip-connection. (c) A block can be split into $b$ branches. }
\label{fig:block_ResNet152}
\end{figure}

We reuse a \rn~\cite{heDeepResidualLearning2016} model pre-trained on ImageNet \cite{russakovskyImageNetLargeScale2015}. This model consists of five groups of blocks, which we call \emph{phases} (\figurename~\ref{fig:arch}): conv1, conv2\_x, conv3\_x, conv4\_x, conv5\_x. The conv1 phase contains a convolutional layer and a max-pooling layer; this can be seen as a special block. All other phases include multiple \emph{bottleneck blocks} (\figurename~\ref{fig:block_ResNet152} (a)).
A bottleneck block (\figurename~\ref{fig:block_ResNet152} (b)) consists of 3 convolutional layers and a skip-connection path (directly connecting input and output). All bottleneck blocks within a phase are identical, with the exception of the first one, which includes an additional convolutional layer in its skip-connection path. Feature map downsampling is performed by conv1, conv3\_1, conv4\_1, and conv5\_1. As seen in \figurename~\ref{fig:arch}, \rn repeats the same blocks many times in each phase.

While our study focuses on \rn, the reduction step described in Section~\ref{sec:reduction_step} can, in principle, also be applied to other ResNet architectures.
The important properties are the skip-connections and that we have phases consisting of identical blocks.
In this case, we can \emph{configure} the phases by deciding upon the number of blocks they contain; the initial \rn serves as a template.

\subsection{Reducing: Model block pruning}

\label{sec:reduction_step}

The ResNet architecture has the benefit of a modular structure consisting of identical blocks with skip-connections. It allows us to perform a neural architecture search that aims to find the optimal number of blocks in each phase. Since the goal here is to \emph{reduce} the network, the maximum number of blocks in each phase is determined by the original \rn architecture. To ease the notation, we will name the architectures as ResNet\_$x_1$\_$x_2$\_$x_3$\_$x_4$, where $x_i$ is the number of blocks used in phase $i$, the conv1 phase is phase $0$. Following this notation, the original \rn is written as ResNet\_3\_8\_36\_3; the smallest possible network in this regime is \rnbase, which contains only one block per phase.

For \rn, the search space consists of $2592$ possible architectures. Extensively evaluating all of them once on one dataset could take several weeks using GPU, which is computationally expensive. Thus, we resort to a greedy forward selection approach to narrow down the search space. This approach, termed forward block selection v1, starts with the simplest architecture \rnbase and successively adds one block at a time until no improvement can be observed. Within this logic, we fill one phase before moving to the next phase until we get ResNet152 (ResNet\_3\_8\_36\_3). The pseudocode of the procedure is given in Appendix~\textit{A} (Algorithm~\ref{algo:reduce2}). This method reuses parameters pre-trained on ImageNet as initialization.

An alternative approach would be to add blocks in a cyclic way \ie always add the block to the phase with the fewest blocks and where blocks may still be added. However, we found in preliminary experiments that these two approaches lead to the same results, so we do not present results for different approaches.

\subsection{Recycling: Ensembling network branches}

\label{sec:recycling}

One way to recycle a given pre-trained backbone is to split a part of it into an \emph{ensemble of branches}. It means, starting from a certain layer, partitioning convolutional filters and adjusting layer connections so that all subsequent layers are influenced by exactly one of the filter sets in the partition. It creates one sub-network (branch) for each filter set in the partition (\figurename~\ref{fig:block_ResNet152} (c)). Each branch ends with a branch-specific output layer with softmax activation. The branches are merged by averaging the probability distributions predicted by different branches. In this sense, the branches can be seen as ensemble members whose votes are aggregated with an average.

We seek to do the branching in such a way that the overall model size and FLOPs (multiply-adds) are preserved. Each branch is constructed by adjusting the number of output channels of convolutional layers. For each branch, the number of output channels $C_b$ is computed as follows:

\begin{equation}
    C_b = \left\lfloor \frac{C_o}{\sqrt{b}} \right\rfloor - a
    \label{eq:channels_per_branch}
\end{equation}

where $C_o$ is the number of output channels in the original pre-trained convolutional layer, $\lfloor \cdot \rfloor$ is the floor function, and $b$ is the number of branches to create. The variable $a$ in Equation~\ref{eq:channels_per_branch} is useful to ensure that the total number of parameters in the branches is reduced or equal to the original model's parameter count. The value of $a$ relies on: the number of branches $b$, the starting point from which to split the model, the number of classes of the downstream task, and the model structure (\eg \rnbase or \rn). $a$ can be empirically computed by finding the smallest value that satisfies the total parameter count constraint. This computation results in each branch having $C_b$ output channels as shown in \figurename~\ref{fig:arch} (``Ours'' columns). Each branch is initialized with a part of pre-trained kernels (pseudocode is given in Appendix~\textit{B} Algorithm~\ref{algo:splittingalgo}).

In our design, we keep the conv1, conv2\_x, and conv3\_x phases as the shared stump (preprocessor), and we create the branches starting from conv4\_1 \ie the first block of the conv4\_x phase. This way, conv4\_x and conv5\_x effectively constitute an ensemble of independent sub-networks (branches). The reason for branching at the last two phases is that these two phases include the most number of parameters and output channels, allowing for more branches to be made. By maintaining the total parameter count, the more branches are created, the fewer parameters each branch will have.

There can be many methods to train this ensemble of branches. One method is to train branches as if they were individual models without using specific ensemble training techniques. We call this training approach {\bf naive ensemble}. Other choices include bagging~\cite{breimanBaggingPredictors1996}, Snapshot Ensemble~\cite{huangSnapshotEnsemblesTrain2017} and Fast Geometric Ensemble~\cite{garipovLossSurfacesMode2018}. All ensemble training methods can be accompanied by random data augmentation to create diverse ensemble members. The stump (conv1, conv2\_x, conv3\_x) shared by branches is frozen; it keeps the pre-trained parameters. 

\section{Experiments}

We carry out comprehensive experiments in this section. We evaluate our methodology's reduction step compared to finer-grained pruning approaches. We perform model selection and evaluation for our methodology's reduction and recycling steps. Finally, we demonstrate the performance of the proposed model on a large benchmark. We also report results from a multi-criteria comparison, including accuracy, inference time, and the number of parameters.

\subsection{Datasets}

We conduct experiments on Meta-Album micro~\cite{ullahMetaalbumMultidomainMetadataset2022}, which is a large image classification benchmark of $40$ datasets. These datasets come from various domains such as ecology (fauna and flora), manufacturing (textures, vehicles), human actions, and optical character recognition. The datasets contain images with different scales, ranging from microscopic to human-sized, to images taken from remote sensing. These images represent a diverse and challenging set of data for our models to learn. Each dataset contains $128 \times 128$ images, and the goal is to classify each image into one of $20$ different classes. Each class in each dataset has $20$ images for the training and $20$ images for the testing (\ie both the training set and the test set contain $400$ images each). The choice of using this benchmark for our experiments is due to its diversity of problems and because it demonstrates low-resource scenarios (\ie few training examples available). These scenarios are common in real-world applications and highlight challenges faced when developing learning models with limited data. 

\subsection{Implementation Details}

We formatted an extra held-out dataset to perform model selection and hyper-parameter validation: we carved out a dataset from an OCR dataset of images of alphanumeric characters in the wild~\cite{lucasICDAR2003Robust2003}. This dataset is dimensioned similarly to the Meta-Album benchmark. It has $20$ classes, $15$ images per class for training, $5$ images per class for validation, and $20$ images per class for the test. We call this dataset ICDAR-micro (named after its original source).

In the spirit of Automated Machine Learning, we optimize all the hyper-parameters of our model with the ICDAR-micro dataset.\footnote{Admittedly, we could have used more datasets, but, as it turns out, we already obtained quite good results with this strategy.} This allows us to use a simple train/test split when we evaluate our method on the Meta-Album benchmark since we do not need validation splits to select any setting. The identified hyper-parameter values are summarized in Appendix~\textit{C}.

We also used the ICDAR-micro dataset to tune the reduction and recycling procedure for our low-resource regime: (1) how much pruning of the original backbone we could do (to reduce inference time and storage); (2) into how many branches we should split the pre-trained backbone (to gain performance by ensembling). These experiments are detailed in Section~\ref{sec:prelim_pruning} and \ref{sec:prelim_ensembling}.

All experimental results are averaged over repeated runs. 95\% confidence intervals are computed over repeated runs using t-distribution.

\subsection{Results from the reduction procedure} 
\label{sec:prelim}

In this section, we summarize two results observed in the reduction procedure. In Section~\ref{sec:pruning_granularity_study}, we compare the inference speed of our methodology's {\em reduction} step with finer-grained pruning approaches that could be applied to reduce the network. Second, we study the extent to which we could reduce the network without significantly degrading performance on the ICDAR-micro dataset in Section~\ref{sec:prelim_pruning}.

\subsubsection{Comparing pruning granularities}

\label{sec:pruning_granularity_study}

\begin{figure}
    \centering
    \includegraphics[width=\columnwidth]{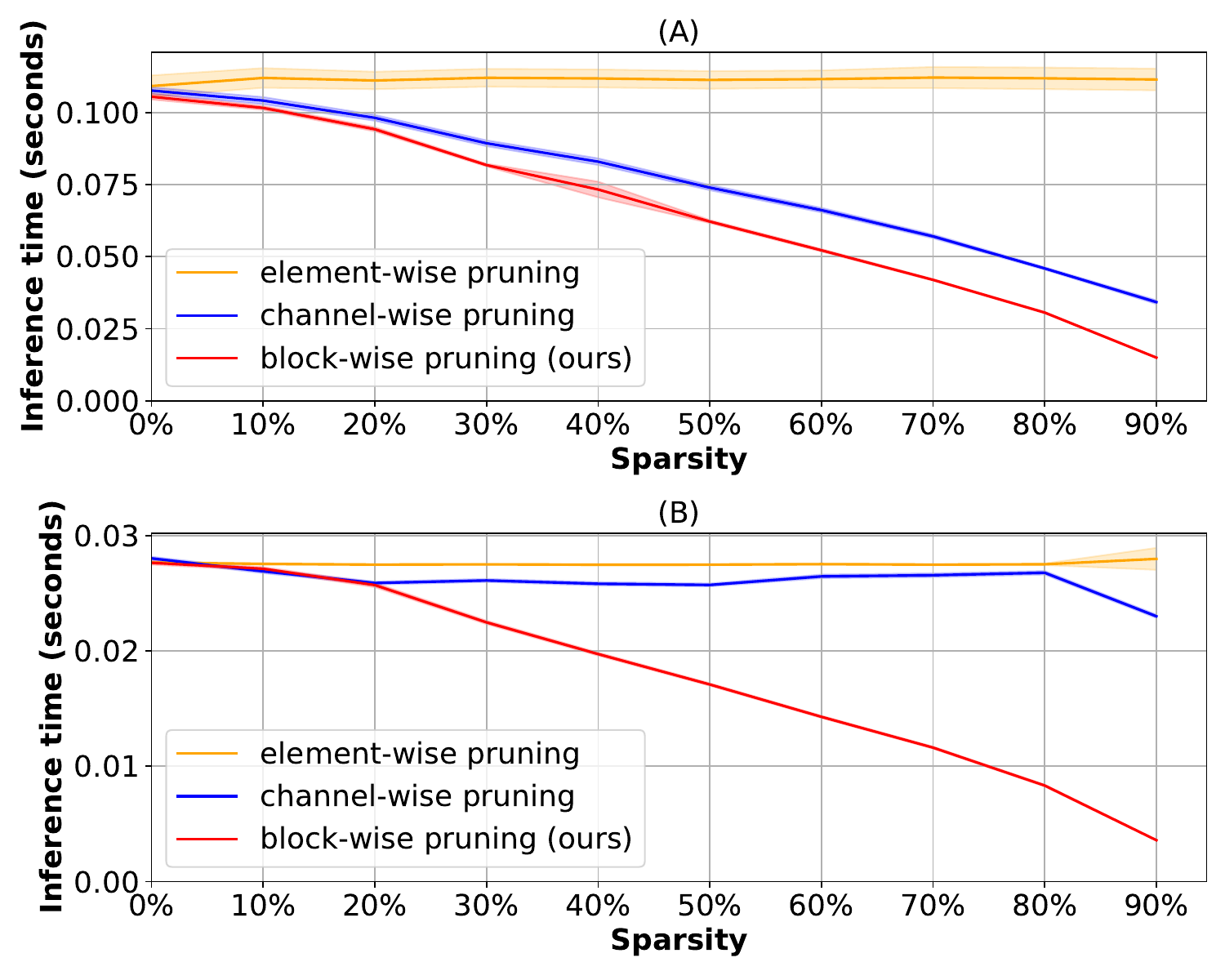}
    \caption{{\bf Inference time of the pruned model.} The upper figure (A) shows the evaluation on CPU; the lower figure (B) shows the evaluation on GPU. Vertical axes show the number of seconds required to run one forward pass for one image; the lower, the better. Horizontal axes denote sparsity. $0\%$ means no pruning, $90\%$ means $90\%$ of parameters are removed. Curves and $95\%$ confidence interval are computed with at least 200 independent runs.}
    \label{fig:pruning_granularity_study_both}
\end{figure}

This section investigates the potential gain in hardware inference speed of different pruning granularities; here, we ignore the classification performance. We compare 3 levels of pruning granularity: (1) element-wise pruning, where individual parameters are removed; (2) channel-wise pruning, where channels in a convolutional layer are removed; (3) block-wise pruning, where one ResNet block (\figurename~\ref{fig:block_ResNet152}, one block contains three layers) is removed at a time; the latter is our approach. Element-wise pruning implementation is the official pruning package of PyTorch~\cite{paszkePyTorchImperativeStyle2019}, where pruned individual parameters are set to zero. Channel-wise pruning is the Torch-Pruning implementation~\cite{fangTorchPruning2022}, where parameter tensors are effectively slimmed to remove pruned channels. Block-wise pruning skips pruned blocks.

We compare all pruning approaches on the same backbone network: ResNet152~\cite{heDeepResidualLearning2016}. To measure the potential gain in hardware inference speed, we report inference time in seconds, both on CPU (Intel Xeon Gold 6126) and GPU (GeForce RTX 2080 Ti), across different sparsity levels (\figurename~\ref{fig:pruning_granularity_study_both}). Sparsity is the fraction of parameters that are removed compared to the full ResNet152. In our experiments, individual parameters or convolutional channels are removed randomly for element-wise pruning and channel-wise pruning, and the sparsity is approximately uniformly distributed among layers; block-wise pruning removes blocks following the block order (Appendix~\textit{A} Algorithm~\ref{algo:reduce2}).

The results show that element-wise pruning (yellow curves in \figurename~\ref{fig:pruning_granularity_study_both}) provides limited speedup, which is expected because the number of parameters does not effectively change, tensor sparsity is not exploited; curve fluctuation corresponds to noise in evaluation. Channel-wise pruning (blue curves in \figurename~\ref{fig:pruning_granularity_study_both}) provides limited speedup on GPU but effectively accelerates the inference on CPU. Our block-wise pruning strategy (red curves in \figurename~\ref{fig:pruning_granularity_study_both}) results in the most significant computational gain on both CPU and GPU, particularly when sparsity increases. The advantage of block-wise pruning over channel-wise pruning and the poor result of channel-wise pruning on GPU can be explained by the use of parallel computing in modern CPUs and GPUs. The computation within each layer is parallelized because it involves matrix multiplications~\cite{chellapillaHighPerformanceConvolutional2006, jiaLearningSemanticImage2014}. The effect of channel-wise pruning boils down to reducing the parameter tensor size within each layer. Its benefit diminishes when there is more parallelism in the hardware processor, which is the case of GPUs~\cite{navarroSurveyParallelComputing2014, kirkNVIDIACUDASoftware2007}. On the other hand, \rn, like many other modern neural networks, consists of sequentially concatenated layers. Computation between layers is blocking; the execution of one layer needs to wait for the result of the previous layer. Removing blocks (hence layers) reduces the number of sequentially blocking computations.

\subsubsection{Reducing the number of blocks}
\label{sec:prelim_pruning}

\begin{figure}
    \centering
    \includegraphics[width=\columnwidth]{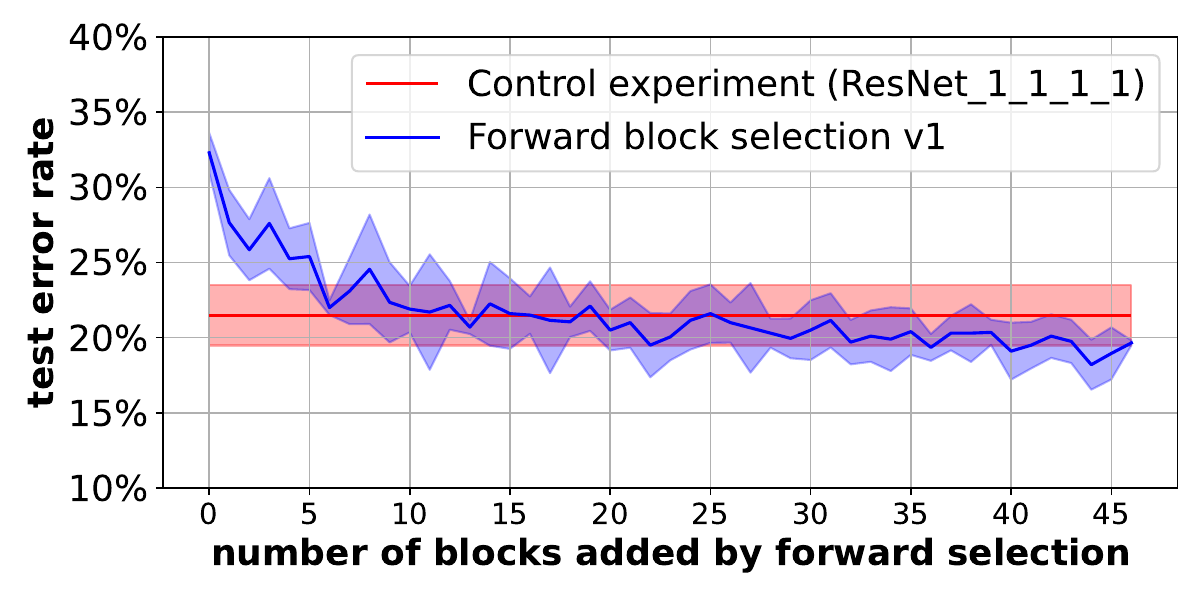}
    \caption{{\bf Optimizing the number of blocks.} Experiments on ICDAR-micro data, applying the forward selection algorithm presented in Appendix~\textit{A} Algorithm~\ref{algo:reduce2} (blue curve). The red horizontal line indicates the final performance of the control experiment. The vertical axis denotes the error rate (1 - accuracy) on the test set; the lower, the better. Curves and 95\% confidence intervals are computed with 5 independent runs (5 random seeds).}
    \label{fig:pruning}
\end{figure}

We ran the forward selection algorithm, presented in Appendix~\textit{A} (Algorithm~\ref{algo:reduce2}), on ICDAR-micro to determine the smallest number of blocks we could keep in our low-resource regime without significantly degrading performance. The results are shown in \figurename~\ref{fig:pruning}. The blue curve is the learning curve of the forward selection algorithm, where each point in the curve is obtained by training the candidate architecture for $300$ epochs. In this plot, the stopping criterion is deactivated, so the forward selection algorithm added $47$ blocks in the end, which corresponds to $14100$ epochs in total. As a control experiment, we compare the performance of the forward selection algorithm with training \rnbase for the same overall budget of epochs; the red horizontal line shows the final performance of training \rnbase for $14100$ epochs. We observe that the final point of the blue curve does not significantly improve upon the red horizontal line.

Hence, these results showed that the model \rnbase is sufficient for the given type of tasks and budget, and adding more blocks does not result in significant improvement. Therefore, the recycling/ensembling technique will be applied to \rnbase in subsequent experiments.

\subsection{Results from the recycling procedure (varying the number of branches)}
\label{sec:prelim_ensembling}

\begin{figure}
    \centering
    \includegraphics[width=\columnwidth]{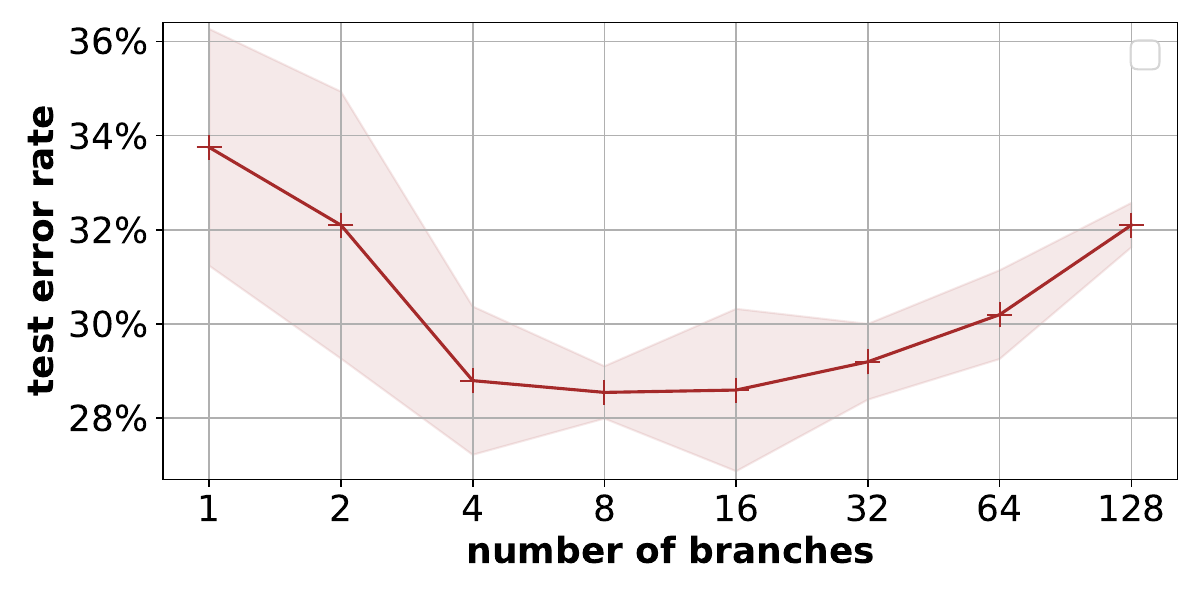}
    \caption{{\bf Optimizing the number of branches.} Experiments on IDCAR-micro data, varying the number of branches. Branches are trained by the naive ensemble approach. In this plot, the number of branches is doubled each time, starting from 1 (\rnbase without splitting) up to 128. The optimum (lowest error rate) is at $8$ branches. Curves and 95\% confidence intervals are computed with 5 independent runs (5 random seeds).}
    \label{fig:branch_opt}
\end{figure}

Since our approach preserves the total parameter count in the Recycling step, the number of branches provides a trade-off between the number of ensemble members and the capacity of each member. We used ICDAR-micro again to determine an optimal number of branches for our low-resource regime. 

In \figurename~\ref{fig:branch_opt}, we vary the number of branches recycled from our backbone (trained by the naive ensemble approach). The results show that ``$8$ branches'' is optimal (performances are quite insensitive to the exact number of branches around that point). In what follows, we use \rnbase with $8$ branches (splitting starts from conv4\_1), denoted as \rnbase-8\_branch or RRR-Net.

We also tried alternative ensembling methods, including bagging~\cite{breimanBaggingPredictors1996}, Snapshot Ensemble~\cite{huangSnapshotEnsemblesTrain2017}, and Fast Geometric Ensemble (FGE)~\cite{garipovLossSurfacesMode2018}, to train branches on the ICDAR-micro data. However, the experiments showed that the naive ensemble approach leads to better performances than the alternative ensembling methods. For this reason, we adopt the naive ensemble approach to train branches.

\begin{figure*}
    \centering
    \includegraphics[width=\textwidth]{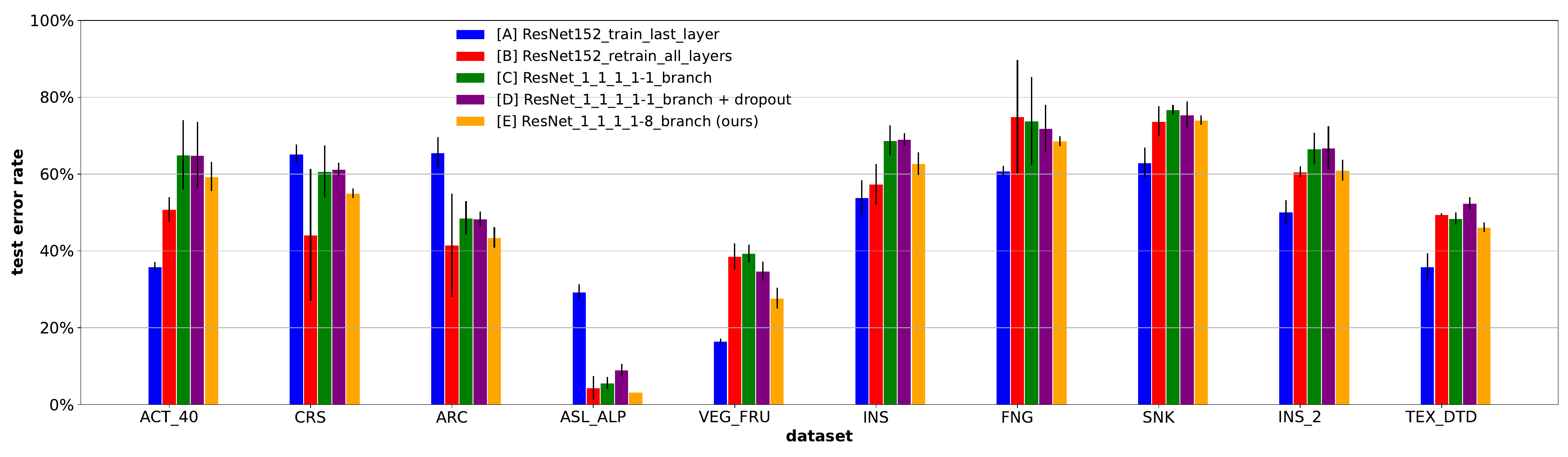}
    \includegraphics[width=\textwidth]{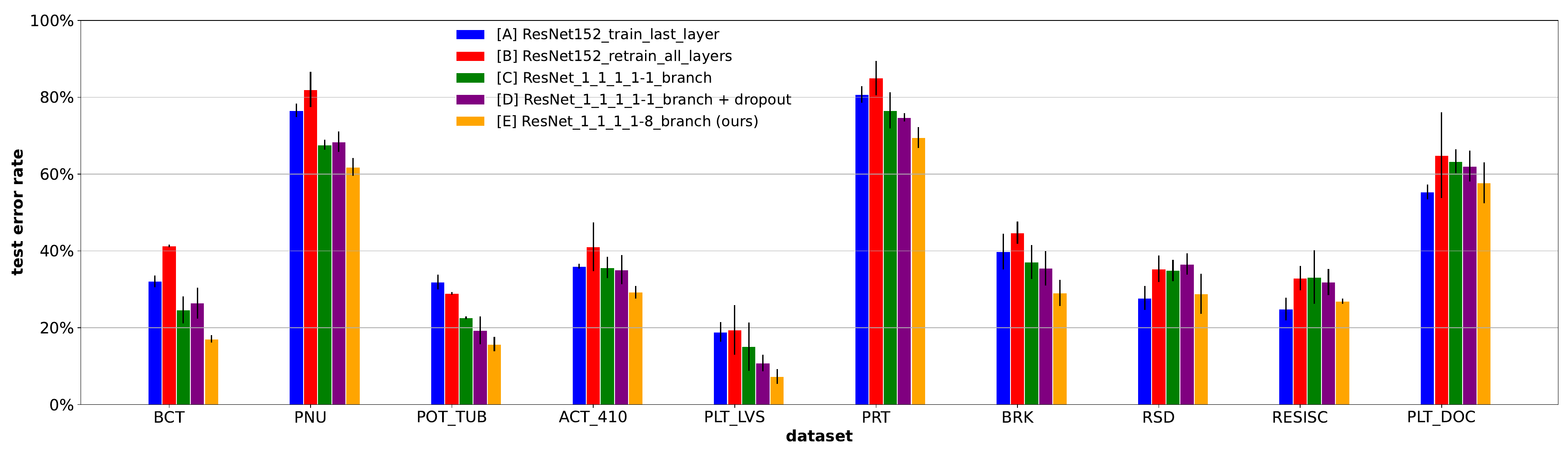}
    \includegraphics[width=\textwidth]{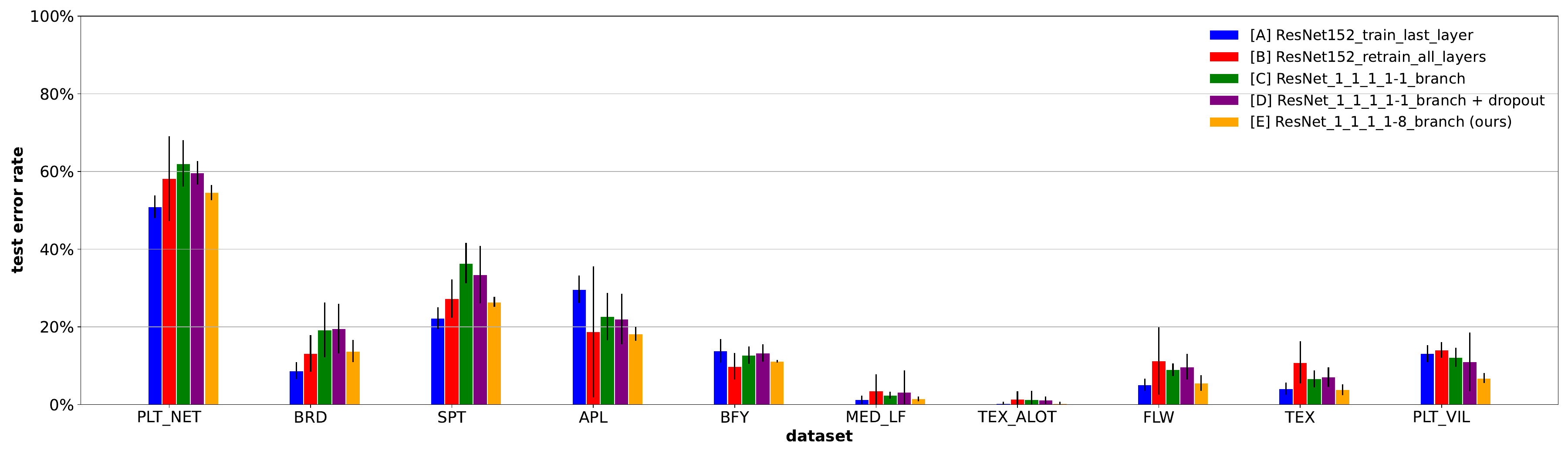}
    \includegraphics[width=\textwidth]{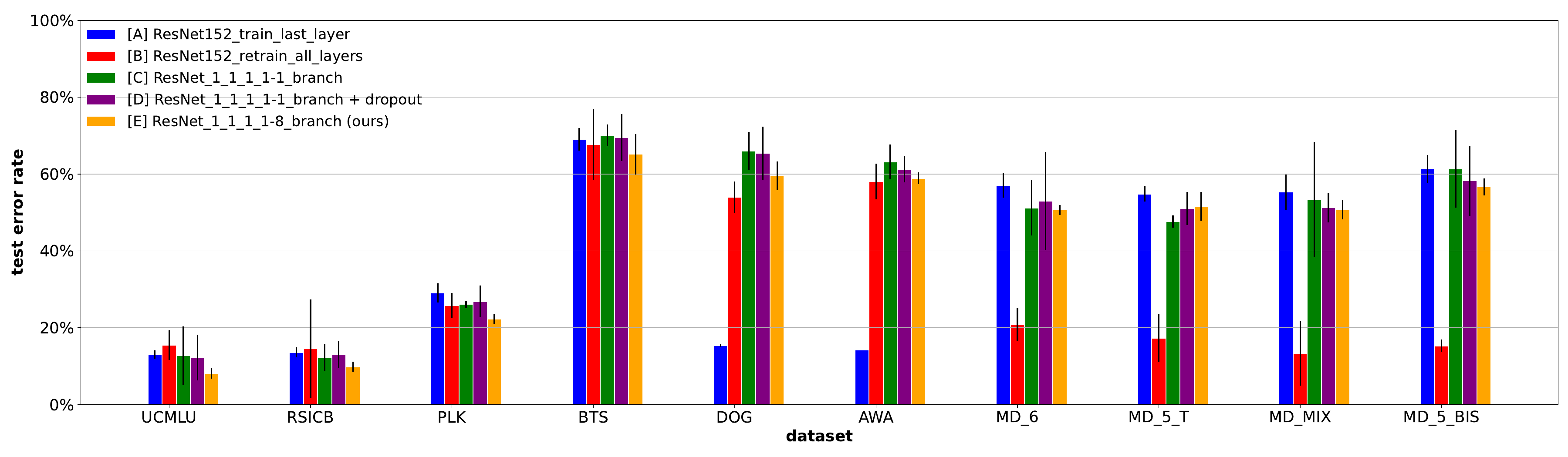}
    \caption{{\bf Results on the $40$ datasets of Meta-Album micro.} The vertical axis is the error rate on the test set; the lower, the better. Five models are shown for each dataset: [A] training the last layer (classification head) of the pre-trained ResNet152; [B] retraining all layers of the pre-trained ResNet152; [C] \rnbase without splitting; [D] \rnbase without splitting but we apply dropout on the last two blocks; [E] \rnbase-8\_branch (RRR-Net) trained by the naive ensemble approach. The 95\% confidence intervals are computed with 3 independent runs (3 random seeds).} 
    \label{fig:main_results}
\end{figure*}

\subsection{Results on the Meta-Album benchmark}
\label{sec:results}

\begin{figure}
    \centering
    \includegraphics[width=\columnwidth]{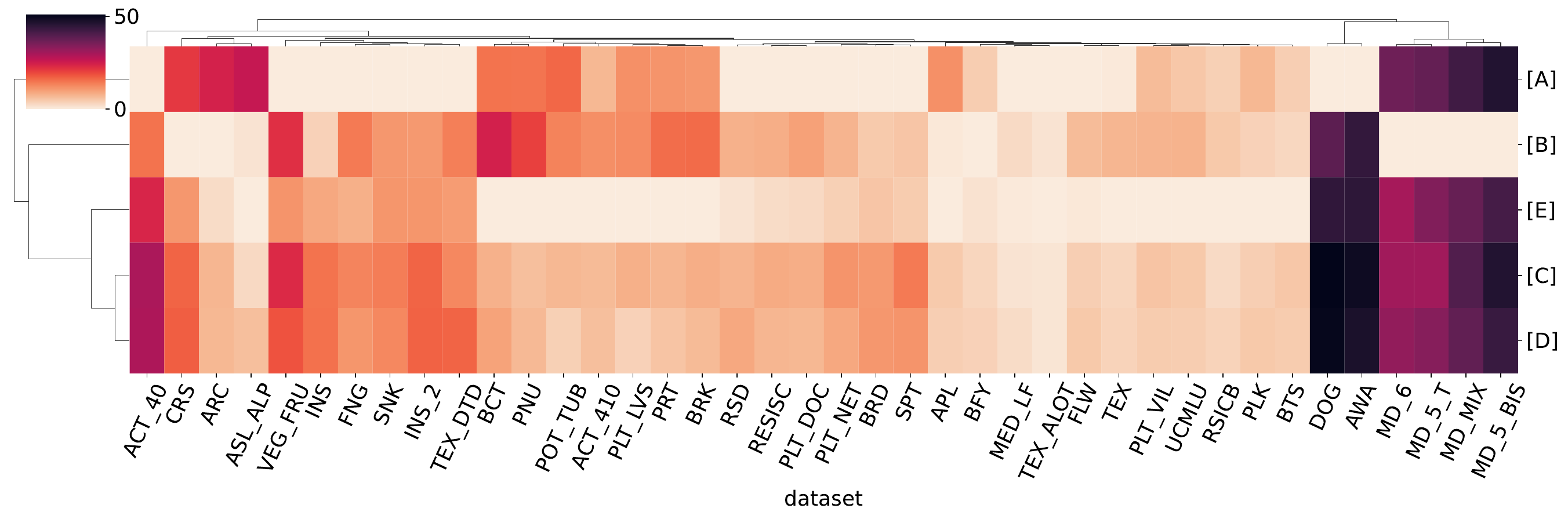}
    \caption{{\bf Hierarchically-clustered heatmap of the model-dataset matrix.} Each column is one dataset. Each row is one model, [E] is our proposed model. The heatmap values are the error rate difference with respect to the best-performing model on the same dataset; the lower, the better. }
    \label{fig:clustermap}
\end{figure}

\begin{figure}
    \centering
    \includegraphics[width=\columnwidth]{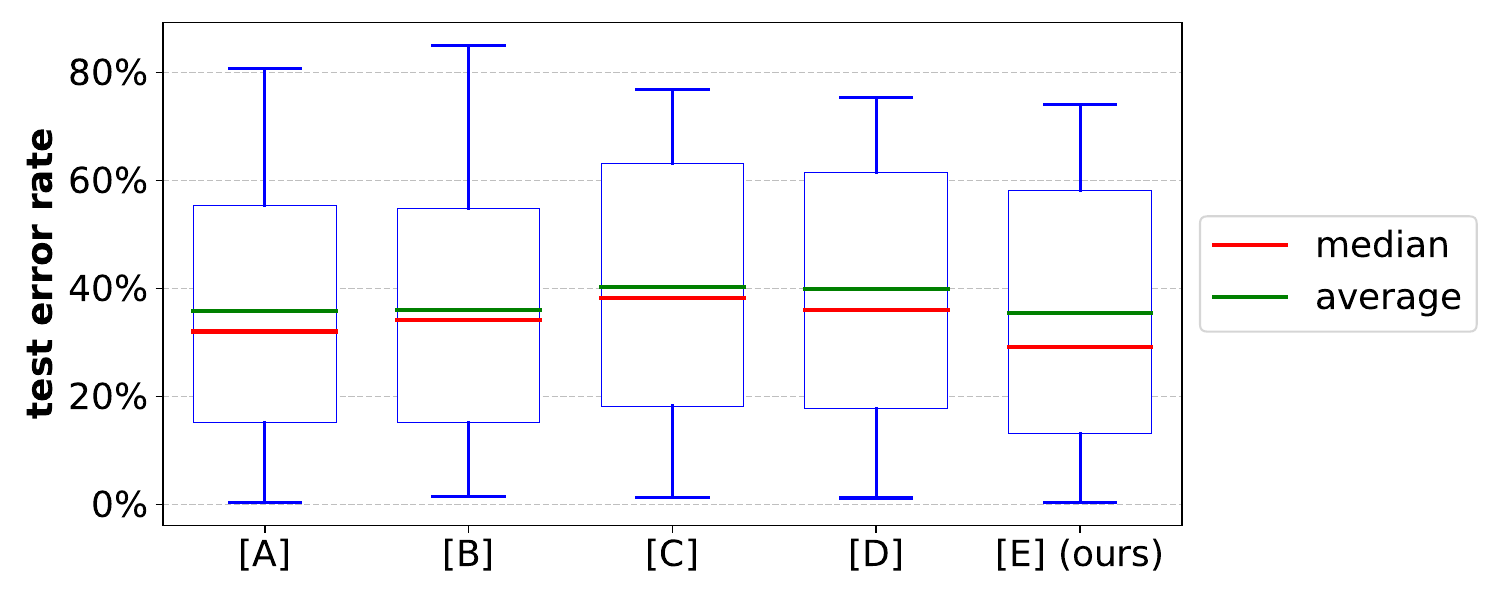}
    \caption{{\bf Box plot of test error rates for each model.} The vertical axis shows the error rate; the lower, the better. Our model [E] has the best average, median, 25\% quantile, and worst-case (upper whiskers) error rates. Baselines [A] and [B] have better 75\% quantiles.}
    \label{fig:boxplot}
\end{figure}

The results on the Meta-Album benchmark~\cite{ullahMetaalbumMultidomainMetadataset2022} are shown in \figurename~\ref{fig:main_results}. We compare 5 models:

{\small 
\begin{itemize}
    \item \textcolor[rgb]{0.13333333333,0,0.99607843137}{[A] blue} Training the last layer of the pre-trained ResNet152 
    \item \textcolor[rgb]{1,0,0}{[B] red} Retraining all layers of the pre-trained ResNet152
    \item \textcolor[rgb]{0.00392156862,0.50196078431,0}{[C] green} \rnbase without splitting
    \item \textcolor[rgb]{0.50196078431,0,0.50196078431}{[D] purple} \rnbase without splitting but we apply dropout~\cite{tompsonEfficientObjectLocalization2015} on the last two blocks
    \item \textcolor[rgb]{1,0.64705882352,0}{[E] yellow} \rnbase-8\_branch (RRR-Net) trained by the naive ensemble approach.
\end{itemize}
}

We use models [A] and [B] as baselines. Depending on the distribution shift between the pre-training dataset (ImageNet) and the target dataset, models [A] and [B] can have different relative rankings. Model [E] is our proposed model using ensembling with 8 branches. Models [C] and [D] are controls for [E]. Model [C] is the reduced network without splitting and ensembling; it only has one branch. Model [D] replaces the recycling step of model [E] with dropout~\cite{tompsonEfficientObjectLocalization2015}: in the last two blocks, we inserted dropout after each activation function (except for the softmax activation). Of several tried dropout rates ($0.05$, $0.1$, $0.2$, $0.3$, $0.4$, $0.5$, $0.6$), we report the most favorable result for model [D] (dropout rate $= 0.05$).

The Meta-Album benchmark~\cite{ullahMetaalbumMultidomainMetadataset2022} is challenging since it covers a wide diversity of domains and scales. As shown in \figurename~\ref{fig:clustermap}, the ranking of different models varies with respect to each dataset. None of the models consistently outperforms all others in all datasets. For example, the overall best baseline model [A] failed on datasets such as ARC and ASL\_ALP. The degree of similarity between the target dataset and the pre-training dataset (ImageNet) affects the models' relative performance. This pattern is particularly visible for datasets of macroscopic animals (DOG, AWA), for which last-layer fine-tuning ([A]) clearly outperforms the other models. Pre-trained features are readily available for these datasets, except for the last layer. On the other hand, the character recognition datasets (MD\_6, MD\_5\_T, MD\_MIX, MD\_5\_BIS) are very dissimilar to ImageNet; retraining all layers ([B]) clearly outperforms the other models in these character recognition datasets. In contrast, the proposed model [E] outperforms the other models on a wide variety of datasets, including datasets that are quite different from ImageNet \eg BCT, PNU, and POT\_TUB from the microscopy domain; BRK, TEX, and TEX\_ALOT from the texture domain; UCMLU and RSICB from the remote sensing domain.

The diversity of the Meta-Album benchmark~\cite{ullahMetaalbumMultidomainMetadataset2022} made us consider the overall performance of the models under examination. We calculated the average error rates of these $5$ models by averaging over the $40$ datasets in Meta-Album. The average error rates for these $5$ models are 35.8\%, 36.0\%, 40.3\%, 39.9\%, 35.5\%. Similarly, the median error rates of these $5$ models are 32.0\%, 34.1\%, 38.2\%, 36.0\%, 29.2\%. Our proposed model [E] is found to have the best average and median error rates, as shown in \figurename~\ref{fig:boxplot}. To further validate these findings, we conducted Wilcoxon signed-rank tests (one-tailed)~\cite{wilcoxonIndividualComparisonsRanking1945}. These statistical tests indicated that model [E] significantly outperforms model [D] with a p-value of $3\times 10^{-12}$ and significantly outperforms model [C] with a p-value of $8\times 10^{-11}$. Model [E] is also found to be significantly better than model [B] with a p-value of $0.036$. On the other hand, model [E] performs comparably well to model [A] with a p-value of $0.16$. These results indicate that the proposed model [E] performs well on this benchmark with $40$ datasets.

\begin{figure}
    \centering
    \includegraphics[width=\columnwidth]{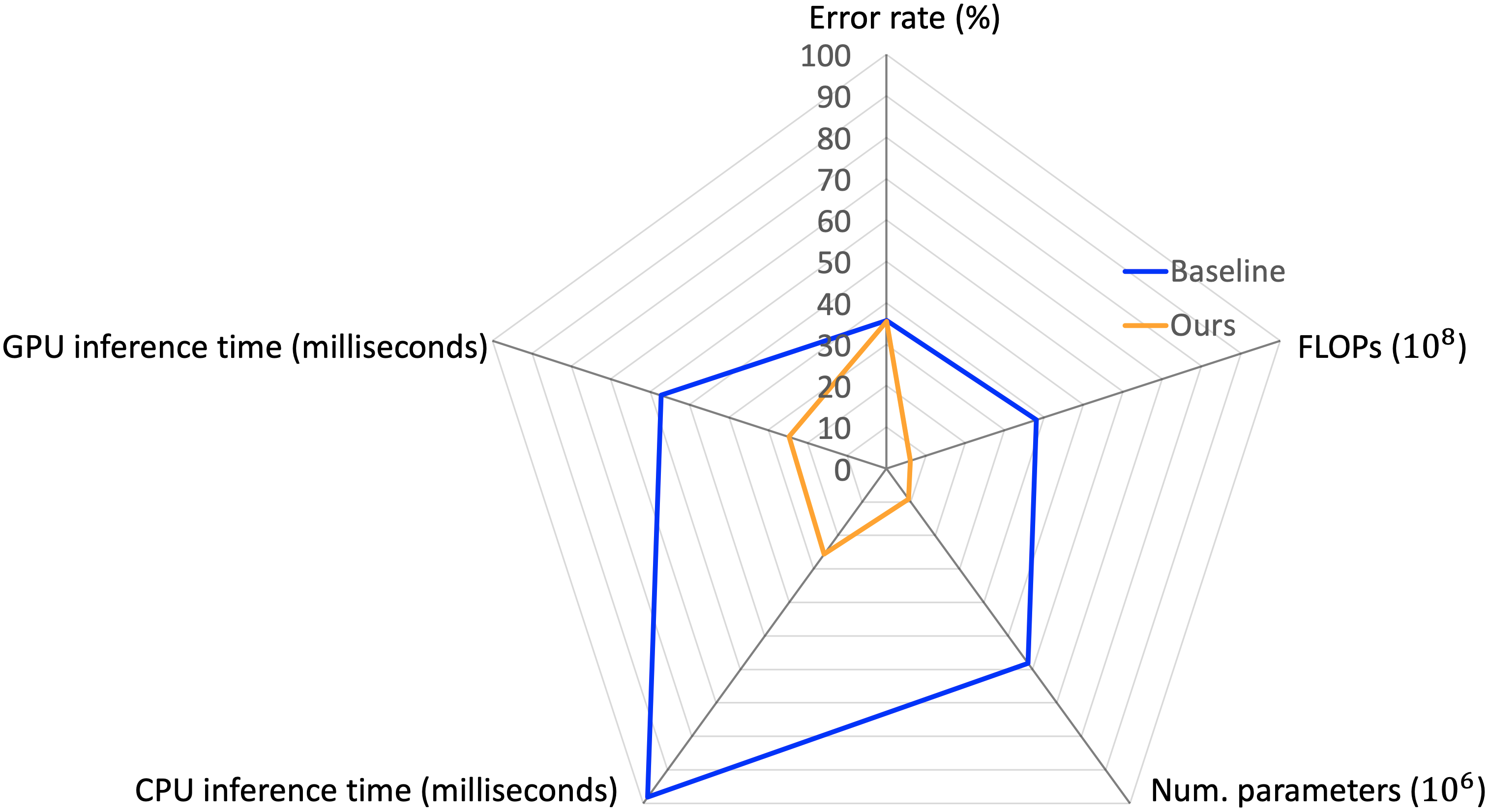}
    \caption{{\bf Multi-criteria comparison.} Our model [E] performs similarly to the overall best baseline [A] while being significantly smaller and faster. Hardware inference time (for one image) is measured with Intel Xeon Gold 6126 and GeForce RTX 2080 Ti.}
    \label{fig:spider}
\end{figure}

Furthermore, we compared the overall best baseline model [A] and our proposed model [E] with respect to five evaluation criteria. These criteria include (1) the average error rate across all $40$ datasets in Meta-Album; (2) the number of parameters; (3) the FLOPs (multiply-adds), which indicate the theoretical inference speed; (4) the inference time on CPU; and (5) the inference time on GPU. We represent the comparison as a spider graph depicted in \figurename~\ref{fig:spider}. Here the baseline error rate (35.8\%) corresponds to the average of the blue bars ([A]) in \figurename~\ref{fig:main_results}, which is the strongest baseline on this benchmark. ``ours'' refers to model [E] \rnbase-8\_branch, whose error rate (35.5\%) corresponds to the average of the yellow bars ([E]) in \figurename~\ref{fig:main_results}. We observe that model [E] uses significantly fewer parameters and FLOPs (multiply-adds) and achieves faster inference speed on both CPU and GPU. In summary, \figurename~\ref{fig:spider} supports the advantage of our model; it indicates that our model ([E]) performs similarly well to the overall best baseline while being significantly more frugal.

\section{Conclusion}

In this paper, we studied the benefits of adjusting pre-trained neural network architectures for image classification in terms of compactness and efficiency. While focusing on the \rn architecture, the paper goes beyond just proposing a model compression technique. It highlights the efficient utilization of pre-trained model knowledge following a ``\textbf{R}euse, \textbf{R}educe, and \textbf{R}ecycle'' principle detailed in Section~\ref{sec:RRR_principle}. We conducted experiments on a large benchmark of $40$ image classification datasets from various domains. We empirically demonstrated that (1) Reducing a pre-trained ResNet backbone to a basic 5-block version yields significant savings in resources (computation and storage) without much loss in classification accuracy, and (2) Recycling the last two blocks by splitting them into multiple branches improves performance, while preserving the same amount of computation and storage. Overall, the resulting model matches the performance of the overall best baseline (if not better) while being smaller and faster. The successful results obtained with the \rn starting backbone suggest that the proposed techniques could be applied to other types of modular architectures. Future work might also include improving the ensemble methods, which could possibly lead to better accuracy. The ensemble nature of the proposed recycling technique lends itself to predictive uncertainty computation.

\section*{Acknowledgments}

We gratefully acknowledge constructive feedback from Dustin Carrión-Ojeda and Romain Egele. This work was supported by the TAU team and the ANR (Agence Nationale de la Recherche, National Agency for Research) under AI chair of excellence HUMANIA, grant number ANR-19-CHIA-0022.

\bibliography{refs}

\newpage 

\section*{Appendix}

\subsection*{A. The block selection Algorithm}

\begin{algorithm}[th]
\caption{The forward block selection v1 algorithm for reducing ResNet.}
\label{algo:reduce2}
\textbf{Input}: ResNet152, Task=\{TrainSet, TestSet\}\\
\textbf{Output}: MiniNet
\begin{algorithmic}[1] %
\STATE \textbf{Initialize} 
\STATE MiniNet = \rnbase + new\_classification\_layer
\STATE MiniNet.train\_all\_layers(TrainSet, epochNum)
\STATE Accuracy$^*$ = MiniNet.evaluate(TestSet)
\STATE OldAccuracy = Accuracy$^*$
\STATE \# Traverse phases
\FOR{$P = 2:5$} 
\STATE \# Traverse blocks
\STATE \# $maxBlock(2) = 3, maxBlock(3) = 8$
\STATE \# $maxBlock(4) = 36, maxBlock(5) = 3$ 
\FOR{$i = 2:maxBlock(P)$} 
\STATE \# Add block $i$ in phase $P$. 
\STATE MiniNet = MiniNet $+ convP\_i$
\STATE \# Continue training, no re-initialization. 
\STATE MiniNet.train\_all\_layers(TrainSet, epochNum)
\STATE Accuracy = MiniNet.evaluate(TestSet)
\IF {(Accuracy - OldAccuracy) / Accuracy$^* \geq \epsilon$ }
\STATE OldAccuracy = Accuracy
\STATE \textbf{continue}
\ELSE
\STATE \textbf{return} MiniNet $- convP\_i$
\ENDIF
\ENDFOR
\ENDFOR
\end{algorithmic}

\end{algorithm}

\subsection*{B. Splitting pre-trained kernels into branches}
\label{sec:appendixB}
\begin{algorithm}[th]
\caption{The algorithm describing how to split pre-trained kernels into branches.}
\label{algo:splittingalgo}
\begin{algorithmic}[1] %
\REQUIRE $P$ the parameter tensors of the pre-trained model
\REQUIRE $N$ denotes the number of branches
\STATE \# Traverse pre-trained layers to split
\FOR{$k = 1 : \text{size}(P)$} 
\STATE \# Traverse input channels at layer $k$
\FOR{$x = 1: C_{b}^{k, in}$} 
\STATE \# Compute output channel indices $I$
\IF {$C_{o}^{k, out} \geqslant (C_{b}^{k, out} \times N)$}
\STATE $I = 1 : (C_{b}^{k, out} \times N)$
\ELSE
\STATE  $I = 1 : C_{o}^{k, out}$
\WHILE{$\text{size}(I) < (C_{b}^{k, out} \times N)$}
\STATE  $I = \text{concatenate}(I, \text{shuffled}(1 : C_{o}^{k, out})\;)$
\ENDWHILE
\STATE $I = I[: (C_{b}^{k, out} \times N)]$
\ENDIF
\STATE $start = 1, end = C_{b}^{k, out}$
\FOR{$i = 1 : N$}
\STATE $T_k[i][:, x, :, :] = P_k[I[start : end], x, :, :]$
\STATE $start = start + C_{b}^{k, out}, end = end + C_{b}^{k, out}$
\ENDFOR
\ENDFOR
\ENDFOR
\end{algorithmic}

\end{algorithm}

We describe in Algorithm~\ref{algo:splittingalgo} the splitting process of pre-trained layers into a set of branches. The input to Algorithm~\ref{algo:splittingalgo} is the parameter tensors $P$ of the pre-trained model and the number of desired branches $N$. The output of this algorithm is the parameter tensors $T$ of the branches. Here tensor indices are assumed to start at 1. At layer $k$, the shape of the 4-dimensional parameter tensor in a convolutional layer is $(C^{k, out}, C^{k, in}, s^{h}, s^{w})$. $C^{k, out}$ means the number of output channels, $C^{k, in}$ means the number of input channels, $s^{h}$ and $s^{w}$ are height and width of convolutional kernels. For the pre-trained model, we add the subscript $o$; for the branch, we add the subscript $b$.

\subsection*{C. Results of Hyper-parameter Tuning}

The cross-entropy loss and the AdamW optimizer have been chosen. The learning rate equals $10^{-3}$; the weight decay equals $0.01$, the batch size equals $32$, and the model parameters are initialized with parameters pre-trained on ImageNet~\cite{russakovskyImageNetLargeScale2015} (as opposed to training from scratch). The number of training epochs for each model is $300$. Except for the pruning experiments, we perform model exponential moving average during the last $60$ epochs for all compared methods (if applicable). Data augmentation includes rotation, translation, scaling, shear, random brightness/contrast/color change, sharpening, image inverting, gaussian noise, motion blur, Jpeg compression, posterization, histogram equalization, and solarization. Two transformations are drawn uniformly at random with replacement and then sequentially applied to each training example.

\end{document}